\documentclass[lettersize,journal]{IEEEtran}
\usepackage{amsmath,amsfonts}
\usepackage{algorithmic}
\usepackage{algorithm}
\usepackage{array}
\usepackage[caption=false,font=normalsize,labelfont=sf,textfont=sf]{subfig}
\usepackage{textcomp}
\usepackage{stfloats}
\usepackage{url}
\usepackage{verbatim}
\usepackage{graphicx}
\usepackage{cite}
\usepackage{booktabs}
\usepackage{multirow}
\usepackage{graphicx}

\usepackage{amssymb}
\usepackage{amsfonts}
\PassOptionsToPackage{prologue,dvipsnames}{xcolor}
\usepackage{colortbl}     
\definecolor{bestred}{RGB}{255,200,200}    
\definecolor{subyellow}{RGB}{255,255,200} 

\usepackage[table]{xcolor}

\makeatletter

\newcommand{\Rmnum}[1]{\expandafter\@slowromancap\romannumeral #1@}
\makeatother

\begin{document}

\title{TraGraph-GS: Trajectory Graph-based Gaussian Splatting for Arbitrary Large-Scale Scene Rendering}

\author{
Xiaohan~Zhang$^{\dagger}$,
Sitong~Wang$^{\dagger}$,
Yushen~Yan,
Yi~Yang,
Mingda~Xu,
and~Qi~Liu*, \IEEEmembership{Senior Member, IEEE}
\thanks{

$^{\dagger}$ indicates equal contribution.

* indicates the corresponding author.

The authors are with the School of Future Technology, South China University of Technology, Guangzhou 511442, China.

E-mail: \{ftxiaohanzhang, ftkanewang, 202364871411, ftyy, ftxumingda\}@mail.scut.edu.cn, drliuqi@scut.edu.cn}
}

\markboth{Journal of \LaTeX\ Class Files,~Vol.~14, No.~8, August~2021}%
{Shell \MakeLowercase{\textit{et al.}}: A Sample Article Using IEEEtran.cls for IEEE Journals}

\IEEEpubid{0000--0000/00\$00.00~\copyright~2021 IEEE}

\maketitle

\begin{abstract}
High-quality novel view synthesis for large-scale scenes presents a challenging dilemma in 3D computer vision. Existing methods typically partition large scenes into multiple regions, reconstruct a 3D representation using Gaussian splatting for each region, and eventually merge them for novel view rendering. They can accurately render specific scenes, yet they do not generalize effectively for two reasons: (1) rigid spatial partition techniques struggle with arbitrary camera trajectories, and (2) the merging of regions results in Gaussian overlap to distort texture details. To address these challenges, we propose TraGraph-GS, leveraging a trajectory graph to enable high-precision rendering for arbitrarily large-scale scenes. We present a spatial partitioning method for large-scale scenes based on graphs, which incorporates a regularization constraint to enhance the rendering of textures and distant objects, as well as a progressive rendering strategy to mitigate artifacts caused by Gaussian overlap. Experimental results demonstrate its superior performance both on four aerial and four ground datasets and highlight its remarkable efficiency: our method achieves an average improvement of 1.86 dB in PSNR on aerial datasets and 1.62 dB on ground datasets compared to state-of-the-art approaches.
\end{abstract}

\begin{IEEEkeywords}
Novel View Synthesis, Arbitrary Large Scenes, 3D Gaussian Splatting, Trajectory Graph.
\end{IEEEkeywords}

\section{Introduction}


\begin{figure}[!t] 
    \centering
    \includegraphics[width=0.5\textwidth]{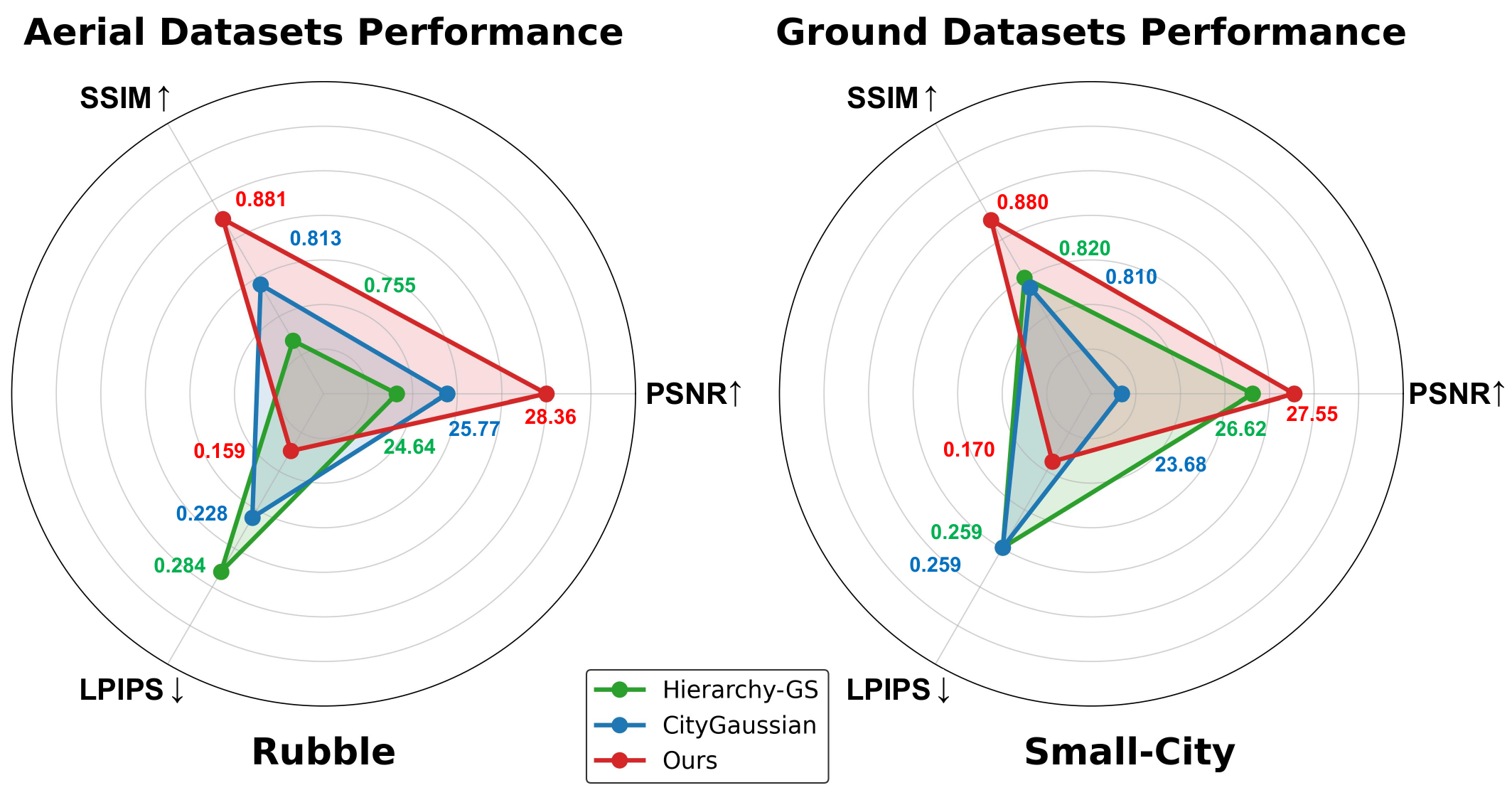} 
    \caption{Existing methods have shown superior performance in specific scenarios but are unsatisfactory to generalize across arbitrary camera trajectories. CityGaussian \cite{liu2024citygaussian} and Hierarchy-GS \cite{kerbl2024hierarchical} have demonstrated the state-of-the-art (SOTA) efficacy on aerial and ground datasets, respectively. CityGaussian exhibits notable restrictions in ground-level data, whereas Hierarchy-GS faces challenges in achieving comparable performance in aerial settings. In contrast, our approach is capable of seamlessly adapting to any camera trajectory and demonstrates superior performance on both aerial and ground datasets.}
    \label{radarplot}
\end{figure}

Novel view synthesis has witnessed remarkable progress in recent years, evolving through several paradigm shifts. Previous approaches relied on geometry-based methods \cite{debevec2023modeling} and light field rendering techniques \cite{buehler2001unstructured}, which either struggled with scene complexity or required dense image sampling. The introduction of Neural Radiance Fields (NeRF) \cite{mildenhall2021nerf} revolutionizes the field by representing scenes as continuous volumetric functions modeled by MLPs. NeRF has demonstrated impressive results for bounded scenes, inspiring numerous follow-up works that improve anti-aliasing \cite{barron2021mip}, enable unbounded scene modeling \cite{barron2022mip}, and accelerate training and rendering through various optimization techniques \cite{muller2022instant} \cite{chen2022tensorf} \cite{reiser2021kilonerf}. Despite these advancements, neural field methods still face challenges with memory limitations and computational demands. More recently, 3D Gaussian Splatting (3DGS) \cite{kerbl20233d} provides an alternative explicit representation using 3D Gaussian primitives with efficient rasterization, achieving both high visual quality and real-time rendering. Various methods have since extended 3DGS to achieve high-precision rendering for small objects and indoor scenes \cite{yu2024mip} \cite{lu2024scaffold}.

For large-scale scenes, processing all data simultaneously requires enormous GPU memory, making divide-and-conquer approaches particularly promising. These methods \cite{liu2024citygaussian} \cite{lin2024vastgaussian} \cite{kerbl2024hierarchical} \cite{zhang2025toy} \cite{zhang2024aerial} supported parallel processing of individual chunks followed by consolidation, which are capable of handling scenes spanning several kilometers with thousands of input images with preserved real-time rendering performance. \textit{Large-scale scenes encompass both large-scale aerial and ground scenes. However, achieving high-quality novel view synthesis on both types of datasets simultaneously presents a significant challenge.} Due to the different camera trajectories in aerial and ground datasets, the spatial partitioning approaches of previous large-scale scene rendering methods are not generally applicable to both types. 
\IEEEpubidadjcol
The division of cameras and sparse point clouds produced by Structure-from-Motion (SfM) into multiple regions, known as spatial partitioning, significantly impacts the quality of scene rendering. Aerial datasets are typically captured with uniform, ideal camera paths, whereas ground datasets suffer from non-uniform trajectories due to road constraints. Moreover, aerial datasets focus on distant objects, whereas ground datasets prioritize near-field details. Consequently, previous spatial partitioning strategies relying on sparse point clouds or camera distribution lack the generalization capability across these diverse dataset types. Furthermore, when merging local representations, these methods are susceptible to generating floating artifacts in the rendered output because of the overlapping of Gaussians across different regions. 

To address these limitations, we propose TraGraph-GS, a trajectory graph-based Gaussian splatting method for high-precision rendering of both aerial and ground scenes. Fig. \ref{radarplot} illustrates the rendering results on various datasets across different scene types, where our method consistently achieves the state-of-the-art (SOTA) performance. Our approach constructs an image connectivity graph where images serve as vertices and feature matching counts as edge weights. It then applies graph segmentation algorithms to partition cameras and sparse point clouds into regions. This visibility-based partitioning strategy adapts to arbitrary camera trajectories, ensuring consistent reconstruction quality. We leverage the connectivity properties of the graph to apply multi-view constraints to disordered images, enhancing texture detail rendering, and design multi-scale Gaussians to improve distant scene rendering. Ultimately, we employ a progressive rendering approach that prioritizes regions by their contribution to novel view synthesis. In detail, primary regions handle main rendering tasks while auxiliary regions dynamically support areas with insufficient coverage. Our contributions are summarized as follows:

\begin{itemize}
\item We propose a trajectory graph-based spatial partitioning method to adaptively divide scenes into various sub-regions, which generalizes well across arbitrarily large-scale scenes with diverse camera trajectories.
\item For disordered image collections, our trajectory graph identifies neighboring images for each viewpoint, enabling us to design effective multi-view constraints and multi-scale Gaussian representations to improve both fine texture details and distant scene rendering quality.
\item A progressive rendering strategy is introduced to avoid the floating points typically caused by the merging of local Gaussians, resulting in more coherent and visually pleasing renderings.
\end{itemize}

\section{Related Work}

\begin{figure*}[!t]
\centering
\includegraphics[width=7.1in]{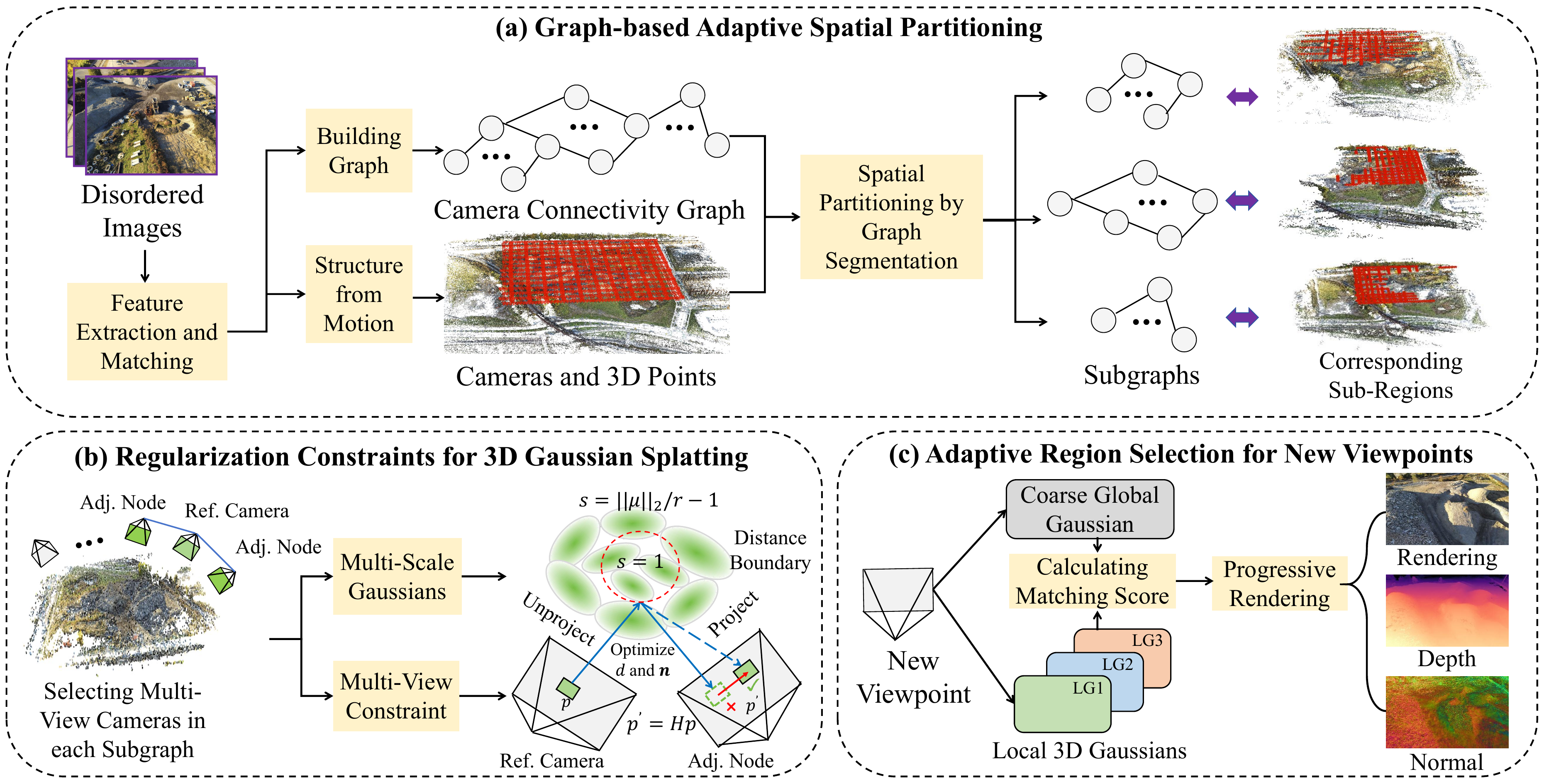}
\caption{Pipeline of our TraGraph-GS: a) Scene Partitioning: From unordered images, we first extract features and construct a camera connectivity graph (images as vertices, feature matching counts as edges weights). Simultaneously, Structure-from-Motion (SfM) provides camera poses and sparse 3D points. Based on this spatial structure, both cameras and points are then partitioned into multiple regions using graph methods. b) Local Training: Building on this partitioning, each region trains local Gaussians independently. During training, neighborhood cameras are identified through subgraphs, where multi-view constraints enhance texture details and multi-scale Gaussians improve distant scene rendering quality. c) Adaptive Rendering: For new viewpoint synthesis, we compute matching scores between global and local Gaussians, thereby ranking them by contribution. Accordingly, high-contribution local Gaussians render primary areas, while low-contribution ones supplement insufficient regions. }
\label{pipeline}
\end{figure*}

\subsection{Novel View Synthesis}

Novel view synthesis has undergone significant advancement through multiple evolutionary phases. Traditional geometry-based methodologies \cite{debevec2023modeling} \cite{seitz1996view} faced limitations when handling complex scenes. Although light field rendering \cite{buehler2001unstructured} provided an alternative framework, it required a large image sampling density. The emergence of NeRF \cite{mildenhall2021nerf} marked a transformative breakthrough by utilizing continuous volumetric representations via MLPs. Achieving compelling performance on constrained scenes, NeRF has catalyzed extensive subsequent research developments. Mip-NeRF \cite{barron2021mip} and Mip-NeRF 360 \cite{barron2022mip} improved anti-aliasing and enabled unbounded scene modeling. NeRF++ \cite{zhang2020nerf++} lifted constraints of bounded scenes, while NeRF in the Wild \cite{martin2021nerf} addressed unconstrained photo collections with varying illumination. Performance optimization became a critical focus, with methods like InstantNGP \cite{muller2022instant}, TensoRF \cite{chen2022tensorf}, and Plenoxels \cite{fridovich2022plenoxels} significantly accelerating training and rendering through multiresolution hash encoding, tensor decomposition, and sparse voxel-based representations, respectively. NSVF \cite{liu2020neural} and KiloNeRF \cite{reiser2021kilonerf} improved efficiency using octree structures and small MLP networks, while K-Planes \cite{fridovich2023k} and Tri-MipRF \cite{hu2023tri} employed multiplane arrangements to represent complex scenes efficiently. Zhang et al. introduced GiganticNVS \cite{Wang2024GNVS}, which focused on large-scale neural rendering of gigapixel-level scenes using an implicit meta-deformed manifold. Liu et al. proposed EditableNeRF \cite{Zheng2024ENERF}, which enabled editing of topologically varying neural radiance fields by using key points. LTM-NeRF \cite{Huang2024LTM} incorporated a Camera Response Function (CRF) module and a Neural Exposure Field to enable HDR view synthesis and 3D local tone mapping from multi-exposure LDR inputs. VD-NeRF \cite{Wu2025VD} decoupled geometry, appearance, and lighting for view-consistent editing and high-frequency relighting. 

Despite these advancements, neural field methods still face challenges with extensive scenes due to memory limitations and computational demands. More recently, 3DGS \cite{kerbl20233d} provided an alternative explicit representation using 3D Gaussian primitives with efficient rasterization, achieving both high visual quality and real-time rendering. Various methods have since extended 3DGS, such as LightGaussian \cite{fan2024lightgaussian}, Compact 3D Gaussian \cite{lee2024compact}, and Mip-Splatting \cite{yu2024mip}, addressing memory constraints, compression, and anti-aliasing, respectively. DeferredGS \cite{Wu2025Deferred} decoupled texture and lighting in Gaussian splatting using deferred shading, enabling editable and realistic rendering results.

\subsection{Large-Scale Ground Scene Rendering}

For large-scale ground scene rendering, divide-and-conquer approaches that partition scenes into chunks have shown significant promise. As these methods enable parallel processing of individual chunks followed by consolidation, they show excellent performance on handling large-scale scenes across kilometers with thousands of input images. For example, Block-NeRF \cite{tancik2022block} partitioned the scene into parts with 50\% overlap and computed a representation for each part. Progressive propagation techniques have proven effective for improving geometric accuracy in ground scene rendering. These approaches leveraged patch matching and multi-view consistency checks to propagate accurate geometric information from well-modeled regions to under-modeled regions \cite{cheng2024gaussianpro}. By incorporating planar constraints and geometric filtering, they produced more compact and accurate scene representations, particularly beneficial for the texture-less regions commonly found in ground scenes. Level-of-detail (LOD) management techniques are essential for efficiently rendering massive ground scenes. Hierarchical approaches construct tree-based structures with interior and leaf nodes, defining efficient methods to merge primitives and determine appropriate cuts through the hierarchy based on projected screen size \cite{luebke2002level}. LOD variants were proposed for point-based representations \cite{dachsbacher2003sequential} \cite{rusinkiewicz2000qsplat}, but these methods cannot be rasterized. Kerbl et al. \cite{kerbl2024hierarchical} introduced a hierarchical 3D Gaussian representation that preserved visual quality for large scenes while providing an efficient LOD solution for rendering distant content. Their approach independently trained large scene chunks and consolidated them into an optimizable hierarchy to improve visual quality, enabling real-time rendering of scenes covering trajectories up to several kilometers with tens of thousands of images. Techniques focused on anti-aliasing at different scales \cite{yan2024multi} further enhanced ground scene rendering quality across varying viewing distances.

\subsection{Large-Scale Aerial Scene Rendering}

Large-scale aerial scene reconstruction has been an active research area for decades. Traditional photogrammetry methods employed Structure-from-Motion (SfM) and keypoint matching for sparse reconstruction, exemplified by foundational systems like Photo Tourism \cite{snavely2006photo} and ``Building Rome in a Day" \cite{agarwal2011building}. These approaches established the groundwork for large-scale reconstruction but struggled with rendering quality and efficiency. For large-scale aerial reconstruction, the divide-and-conquer strategy has proven effective. Block-based methods \cite{turki2022mega} \cite{zhenxing2022switch} divide scenes into spatially adjacent blocks, enabling distributed computing and overcoming memory limitations. These methods typically used geographic boundaries to split scenes, and each block was trained independently and then merged. To address appearance variations common in aerial imagery, several methods incorporated appearance encoding \cite{rematas2022urban}. These approaches modeled lighting changes and exposure differences across captures, reducing reconstruction artifacts. For city-scale aerial rendering, methods like CityGaussian \cite{liu2024citygaussian} employed LoD techniques that dynamically adjusted rendering quality based on viewing distance, critical for maintaining real-time performance across vastly different scales. VastGaussian \cite{lin2024vastgaussian} introduced progressive partitioning for large scenes and decoupled appearance modeling to handle lighting variations in aerial data.

\section{Method}

In this paper, we propose TraGraph-GS, a high-quality rendering method for arbitrary scenes. The pipeline of our method is shown in Fig.\ref{pipeline}. Given disordered scene photographs as input, we first design a graph-based adaptive spatial partitioning approach (Sec.\Rmnum{3}.A) which constructs an image connectivity graph through feature matching and utilizes graph segmentation for adaptive spatial partitioning. Another benefit of using a graph structure is establishing connection relationships between disordered images. We then employ Regularization Constraints for 3D Gaussian Splatting (Sec.\Rmnum{3}.B) to enhance texture details and improve the rendering of distant scenes. Finally, we apply the adaptive region selection for new viewpoints (Sec.\Rmnum{3}.C) to avoid floating artifacts caused by Gaussian overlap in previous methods.

\subsection{Graph-based Adaptive Spatial Partitioning}

\textbf{Building Camara Connectivity Graph}. Previous methods \cite{liu2024citygaussian} \cite{lin2024vastgaussian} \cite{chen2024dogs} merely used point clouds and camera poses obtained through feature extraction and matching \cite{lowe2004distinctive}, and SfM \cite{schonberger2016structure} to divide scenes into multiple regions, while neglecting the structural information of the feature correspondences between images. We fully leverage feature matching information by constructing a camera connectivity graph where images serve as vertices and feature matching counts determine edge weights. The camera connectivity graph is defined as $G = (V, E, w)$, where $V = \{v_1, v_2, ..., v_n\}$ is the set of vertices (or nodes), $E = \{(v_i, v_j) | v_i, v_j \in V\}$ is the set of edges connecting pairs of vertices, and $w: E \rightarrow \mathbb{R}$ is a weight function that assigns a real number weight to each edge. Higher edge weights indicate more extensive co-visibility regions between image pairs. 

\begin{algorithm}[H]
\caption{BFS-Based Graph Segmentation.}\label{alg:alg1}
\begin{algorithmic}
\STATE 
\STATE \textbf{Input:} Camara connectivity graph $G(V,E,w)$, number of segmented regions $k$. 
\STATE \textbf{Definition:} Nodes in each region $\{P_i\}_{i=1}^k$, unassigned nodes $U$.
\STATE \textbf{initialize} $U=V$
\STATE \textbf{for} $i=1,...,k$ \textbf{do}
\STATE \hspace{0.4cm} $P_i=\emptyset$
\STATE \hspace{0.4cm} \textbf{seed node} $n=argmax_{u \in U}(\sum_{(u,v) \in E}w(u,v))$
\STATE \hspace{0.4cm} \textbf{add $n$ to $P_i$}
\STATE \hspace{0.4cm} \textbf{while $|P_i|<\frac{|V|}{k}$ :}
\STATE \hspace{0.8cm} \textbf{adjacent nodes} $N=\{v \in V \backslash P_i|(u.v)\in E,u\in P_i\}$
\STATE \hspace{0.8cm} \textbf{update $P_i=P_i \cup N$, $U=U-N$}
\STATE \textbf{for $u$ in $U$ do}
\STATE \hspace{0.4cm} \textbf{score $s_i(u)=\sum_{v\in P_i}w(u,v)$}
\STATE \hspace{0.4cm} \textbf{score median $m=median(\{s_i(u)_{i=1}^k\})$}
\STATE \hspace{0.4cm} \textbf{high-weight regions $H=\{i|s_i(u)>m\}$}
\STATE \hspace{0.4cm} \textbf{assign nodes $u\in P_i, i=argmin_{i\in H}|P_i|$}
\end{algorithmic}
\label{alg1}
\end{algorithm}

\textbf{Graph Segmentation based on BFS}. Various methods can decompose a graph into multiple subgraphs, such as spectral decomposition-based \cite{fiedler1973algebraic} and Kernighan-Lin-based \cite{kernighan1970efficient} graph segmentation approaches. However, Spectral decomposition exhibits inherent limitations in achieving balanced node distribution, which can lead to potential memory overflow issues. Conversely, Kernighan-Lin algorithms fail to preserve the connectivity of nodes in regions, thereby compromising rendering fidelity. We design a BFS-based graph segmentation method that simultaneously ensures connectivity among nodes within each subgraph while maintaining balanced node distribution across regions. First, we identify the node $n$ with the highest degree among all unassigned nodes and designate it as the seed node. Employ the breadth-first search (BFS) \cite{bundy1984breadth} approach to add nodes until the region contains $|V|/k$ nodes, thus establishing these nodes with edges as a region. Dividing the graph into $k$ regions in this way, subsequently, for each remaining unassigned node, calculate its score $s_i$ relative to each existing region. To maintain balance in the node distribution across regions, assign each unassigned node to the region with the fewest nodes among those regions whose scores exceed the median score for that node. The detailed algorithm is outlined in Algorithm \ref{alg1}.

\subsection{Regularization Constraints for 3DGS}
\textbf{3D Gaussian Splatting}. We implement parallel training of local Gaussians for each region to optimize GPU memory utilization. 3DGS \cite{kerbl20233d} represents scenes using anisotropic 3D Gaussians with learnable attributes. Each Gaussian $G(\mathbf{x})$ is defined as:
\begin{equation}
G(\mathbf{x}) = e^{-\frac{1}{2}(\mathbf{x}-\mathbf{\mu})^T \mathbf{\Sigma}^{-1}(\mathbf{x}-\mathbf{\mu})}
\end{equation}
where $\mathbf{\mu} \in \mathbb{R}^{3 \times 1}$ represents the mean vector (position), and $\mathbf{\Sigma} \in \mathbb{R}^{3 \times 3}$ represents the covariance matrix that defines the shape and orientation of the Gaussian. To ensure the positive semi-definite property of the covariance matrix during optimization, it's typically expressed as $\mathbf{\Sigma} = \mathbf{RSS}^T\mathbf{R}^T$, where the rotation matrix $\mathbf{R} \in \mathbb{R}^{3 \times 3}$ is orthogonal, and the scale matrix $\mathbf{S} \in \mathbb{R}^{3 \times 3}$ is diagonal. For rendering, the color of each pixel $\mathbf{p}$ is calculated by blending $N$ ordered Gaussians $\{G_i| i = 1, ..., N\}$ that overlap $\mathbf{p}$:
\begin{equation}
\mathbf{c}(\mathbf{p}) = \sum_{i=1}^{N} \mathbf{c_i} \alpha_i \prod_{j=1}^{i-1} (1 - \alpha_j)
\end{equation}
where $\alpha_i$ is a learned opacity value, and $\mathbf{c}_i$ is the learned color of Gaussian $G_i$.

\textbf{Multi-View Constraint}. While 3DGS achieves impressive rendering quality, it often struggles with accurate geometry reconstruction, particularly in textureless regions. To address this issue, a multi-view constraint can be introduced based on patch matching techniques. First, for disordered images, we can identify the neighborhood frames of each image from the subgraphs. Second, depth and normal maps are rendered for each image by overlapping the positions and orientations of 3D Gaussians using alpha blending. Then, for each pixel $p$ in a reference view, a spatial plane $(d, \mathbf{n})$ is established where $\mathbf{n} \in \mathbb{R}^{1 \times 3}$ represents the pixel's rendered normal and $d$ indicates the distance from the camera coordinate origin to the spatial plane. For a pixel $\mathbf{p}$ in the reference view, we can use homography warping to project it to its corresponding pixel $\mathbf{p}'$ in a neighborhood view:

\begin{equation}
\mathbf{p}' = \mathbf{Hp}
\end{equation}
in which $\mathbf{H}$ is defined as:
\begin{equation}
\mathbf{H} = \mathbf{K}\left(\mathbf{R} - \frac{\mathbf{tn}^T}{d}\right)\mathbf{K}^{-1}
\end{equation}
$\mathbf{R} \in \mathbb{R}^{3 \times 3}$ and $\mathbf{t} \in \mathbb{R}^{1 \times 3}$ represent the relative rotation and translation between the reference and source views, and $\mathbf{K} \in \mathbb{R}^{3 \times 3}$ is the camera's intrinsic matrix. To mitigate the susceptibility of single-pixel-based scene geometry optimization to influences such as lighting variations, we construct a patch $\mathbf{P}_i$ by selecting several neighboring pixels around each considered pixel \cite{barnes2009patchmatch}. We evaluate the color consistency between corresponding patches using Normalized Cross Correlation (NCC) 
\cite{yoo2009fast} and incorporate this measure into our optimization objective:

\begin{equation}
L_{mv} = \sum_{\mathbf{p} \in \mathbf{P}_i} (1 - \text{NCC}(\mathbf{p}, \mathbf{p}^{\prime}))
\end{equation}
Our final training loss $L$ is defined as:
\begin{equation}
L=(1-\lambda)L_1+\lambda L_{SSIM}+L_{mv}
\end{equation}
where image reconstruction loss functions $L_1$ and $L_{SSIM}$ are illustrated in 3DGS \cite{kerbl20233d}. 

\textbf{Multi-Scale Constraint}. For large-scale scenes with varying distances, standard Gaussian representation often struggles with distant objects. To address this, a position-aware point adaptive control \cite{chen2023periodic} can be employed, applying different scales to Gaussians based on their distance from the viewer:
\begin{equation}
\gamma(\mathbf{\mu}) = \begin{cases}
1 & \text{if } \|\mathbf{\mu}-\mathbf{c}\|_2 < 2r \\
\|\mathbf{\mu}-\mathbf{c}\|_2/r - 1 & \text{if } \|\mathbf{\mu}-\mathbf{c}\|_2 \geq 2r
\end{cases}
\end{equation}
where $r$ denotes the scene radius and $\mathbf{c} \in \mathbb{R}^{3 \times 1}$ denotes the center of cameras. When the directional gradient of a Gaussian ellipsoid exceeds the predetermined threshold, we implement a specific protocol. For cases where the maximum scale of Gaussian ellipsoid $max(s) < g \cdot \gamma {(\mathbf{\mu})}$ is satisfied, we perform a cloning operation on the Gaussian ellipsoid. Conversely, when this criterion is not met, we execute a splitting procedure instead. Additionally, we eliminate Gaussian ellipsoids where $max(s) > b \cdot \gamma {(\mathbf{\mu})}$, with $g$ and $b$ serving as scale threshold parameters that govern these operations. This approach uses larger Gaussians for distant objects and smaller ones for nearby regions, improving the rendering quality across different scales.

\subsection{Adaptive Region Selection for New Viewpoints}

Previous methods \cite{kerbl2024hierarchical} \cite{liu2024citygaussian} \cite{lin2024vastgaussian} for large-scale scene rendering combined all local Gaussians into a global Gaussian to render novel viewpoints. However, these approaches often produced floating artifacts at region boundaries due to Gaussian overlap or create holes from improper pruning of local Gaussians. To address these issues, we propose a progressive rendering strategy that prioritizes local Gaussians with higher contributions to the novel viewpoint, followed by supplementary rendering with lower-contribution Gaussians to refine problematic regions. First, we train a coarse global Gaussian representation to capture the overall scene structure. During this global Gaussian training, we downsample both point clouds and images while avoiding the Gaussian ellipsoid densification process, which prevents excessive memory consumption. Next, we render the novel viewpoint using both local and global Gaussians separately, calculating a matching score $S$ for each local-global Gaussian pair:

\begin{equation}
S=(1-\lambda)L_1+\lambda L_{SSIM}
\end{equation}
$L_1$ and $L_{SSIM}$ are same as Eq. (6). Higher matching scores indicate that the corresponding local Gaussian is more likely to be visible from the novel viewpoint. Then we sort the local Gaussians in descending order $G_1, G_2,...,G_k$ based on their matching scores and calculate the corresponding projection opacity on a pixel $o_1(\mathbf{p}),o_2(\mathbf{p}),,,,,o_k(\mathbf{p})$:
\begin{equation}
o_i(\mathbf{p})=\sum_{i=1}^{N} \alpha_i \prod_{j=1}^{i-1} (1 - \alpha_j)
\end{equation}
where $\alpha_i$ is the opacity of the Gaussian ellipsoid that contributes to the pixel $\mathbf{p}$. Finally, we employ a progressive rendering strategy to compute the rendering of pixel $\mathbf{p}$:
\begin{equation}
\mathbf{c}(\mathbf{p})=\sum_{i=1}^kw_i\mathbf{c}_i(\mathbf{\mathbf{p}})
\end{equation}
in which $\mathbf{c}_i(\mathbf{p})$ is the color of pixel $\mathbf{p}$ rendered by the local Gaussian $G_i$ and the weight of each local Gaussian $w_i$ is defined as:
\begin{equation}
w_i = \begin{cases}
o_i(\mathbf{p}) & \text{if } \sum\limits_{j=1}^{i-1} o_i(\mathbf{p})<\beta \\
0 & \text{if } \sum\limits_{j=1}^{i-1} o_i(\mathbf{p}) \geq \beta
\end{cases}
\end{equation}
where $\beta$ is a hyperparameter, this formula indicates that when the overlapped opacity of local Gaussians with significant contributions at pixel $\mathbf{p}$ is high, it signifies sufficient rendering for this pixel. Conversely, when the overlapped opacity is low, it suggests insufficient rendering of the pixel, necessitating continued rendering with newly obtained local Gaussians.

\section{Experiments}

\subsection{Datasets}

\begin{figure}[!t]
\centering
\includegraphics[width=3.2in]{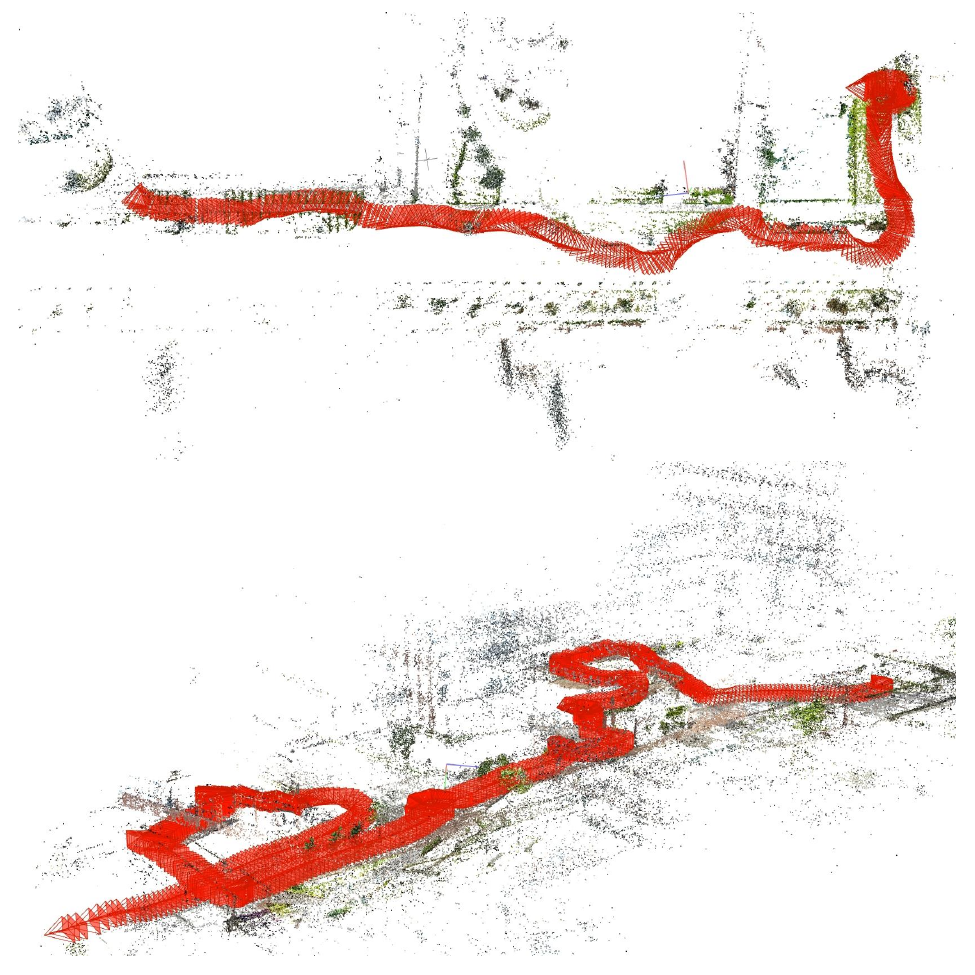}
\caption{Trajectories of our created datasets. The upper figure shows the long and irregular camera trajectory and the sparse point cloud of the $road$ dataset. The lower figure depicts the trajectory and the sparse point cloud of the $Genaral$ dataset, which captures a double-layer building structure.}
\label{track}
\end{figure}

\begin{figure*}[!t]
\centering
\includegraphics[width=7.1in]{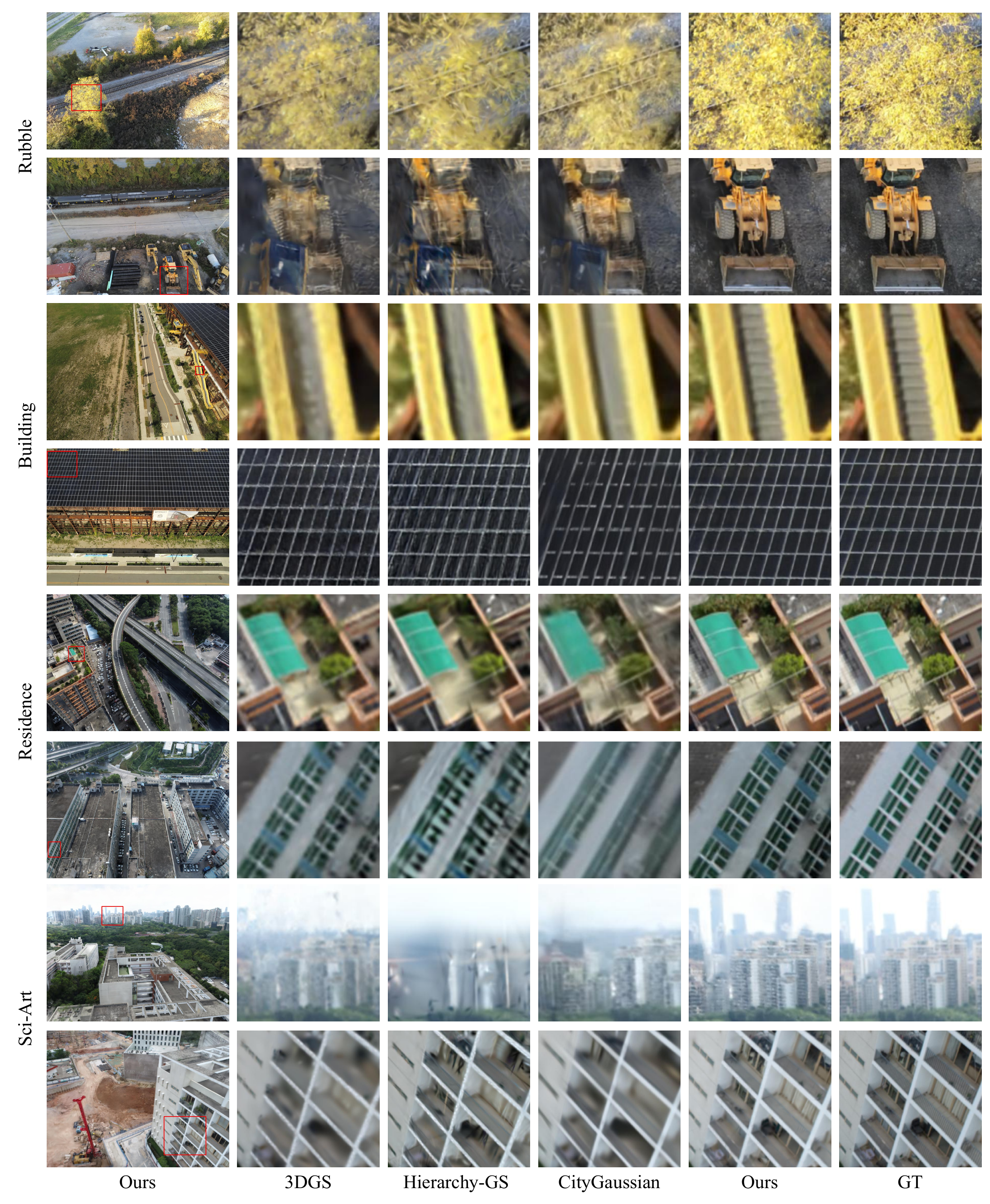}
\caption{Visualization results on four aerial datasets. Aerial scenes present significant challenges in accurately rendering texture details of ground-level objects. In comparison with existing approaches, our method demonstrates superior performance by rendering texture details with enhanced completeness and clarity, while generating vivid representations of distant scenes.}
\label{aerial_fig}
\end{figure*}

\begin{table*}[ht]
    \centering
    \caption{Comparative analysis of neural rendering and Gaussian splatting methods across four benchmark datasets. The table presents quantitative results for PSNR, SSIM, and LPIPS. The best and second best results are highlighted with red shadings in bold and yellow shadings with underline, respectively. The † symbol represents the case where appearance encoding is not employed.}
    \scalebox{1}{
    \begin{tabular}{lcccccccccccc}
        \toprule
        Metrics & \multicolumn{3}{c}{Building} & \multicolumn{3}{c}{Rubble} & \multicolumn{3}{c}{Residence} & \multicolumn{3}{c}{Sci-Art} \\
        \cmidrule(lr){2-4} \cmidrule(lr){5-7} \cmidrule(lr){8-10} \cmidrule(lr){11-13}
        & PSNR$\uparrow$ & SSIM$\uparrow$ & LPIPS$\downarrow$ & PSNR$\uparrow$ & SSIM$\uparrow$ & LPIPS$\downarrow$ & PSNR$\uparrow$ & SSIM$\uparrow$ & LPIPS$\downarrow$ & PSNR$\uparrow$ & SSIM$\uparrow$ & LPIPS$\downarrow$ \\
        \midrule
        
        Mega-NeRF \cite{turki2022mega} & 20.92 & 0.547 & 0.454 & 24.06 & 0.553 & 0.508 & 22.08 & 0.628 & 0.401 & \cellcolor[HTML]{FFFFE0}\underline{25.60} & 0.770 & 0.312 \\
        
        Swich-NeRF \cite{zhenxing2022switch} & 21.54 & 0.579 & 0.397 & 24.31 & 0.562 & 0.478 & \cellcolor[HTML]{FFCCC9}\textbf{22.57} & 0.654 & 0.352 & \cellcolor[HTML]{FFCCC9}\textbf{26.51} & 0.795 & 0.271 \\
        
        3DGS \cite{kerbl20233d} & 20.46 & 0.720 & 0.305 & 25.47 & 0.777 & 0.277 & 21.44 & 0.791 & 0.236 & 21.05 & 0.830 & 0.242 \\
        
        VastGaussian $^\dagger$ \cite{lin2024vastgaussian} & 21.80 & 0.728 & \cellcolor[HTML]{FFFFE0}\underline{0.225} & 25.20 & 0.742 & 0.264 & 21.01 & 0.699 & 0.261 & 22.64 & 0.761 & 0.261 \\
        
        Hierarchy-GS \cite{kerbl2024hierarchical} & 21.52 & 0.723 & 0.297 & 24.64 & 0.755 & 0.284 & 19.97 & 0.705 & 0.297 & 18.28 & 0.590 & 0.316 \\
        
        CityGaussian \cite{liu2024citygaussian} & \cellcolor[HTML]{FFFFE0}\underline{21.55} & \cellcolor[HTML]{FFFFE0}\underline{0.778} & 0.246 & \cellcolor[HTML]{FFFFE0}\underline{25.77} & \cellcolor[HTML]{FFFFE0}\underline{0.813} & \cellcolor[HTML]{FFFFE0}\underline{0.228} & 22.00 & \cellcolor[HTML]{FFCCC9}\textbf{0.813} & \cellcolor[HTML]{FFCCC9}\textbf{0.211} & 21.39 & \cellcolor[HTML]{FFFFE0}\underline{0.837} & \cellcolor[HTML]{FFFFE0}\underline{0.230} \\
        
        Ours  & \cellcolor[HTML]{FFCCC9}\textbf{24.97} & \cellcolor[HTML]{FFCCC9}\textbf{0.838} & \cellcolor[HTML]{FFCCC9}\textbf{0.168} & \cellcolor[HTML]{FFCCC9}\textbf{28.36} & \cellcolor[HTML]{FFCCC9}\textbf{0.881} & \cellcolor[HTML]{FFCCC9}\textbf{0.159} & \cellcolor[HTML]{FFFFE0}\underline{22.14} & \cellcolor[HTML]{FFFFE0}\underline{0.796} & \cellcolor[HTML]{FFFFE0}\underline{0.222} & 22.65 & \cellcolor[HTML]{FFCCC9}\textbf{0.838} & \cellcolor[HTML]{FFCCC9}\textbf{0.197} \\
        \bottomrule
    \end{tabular}
    }
    \label{aerial_tab} 
\end{table*}

\begin{figure*}[!t]
\centering
\includegraphics[width=7.1in]{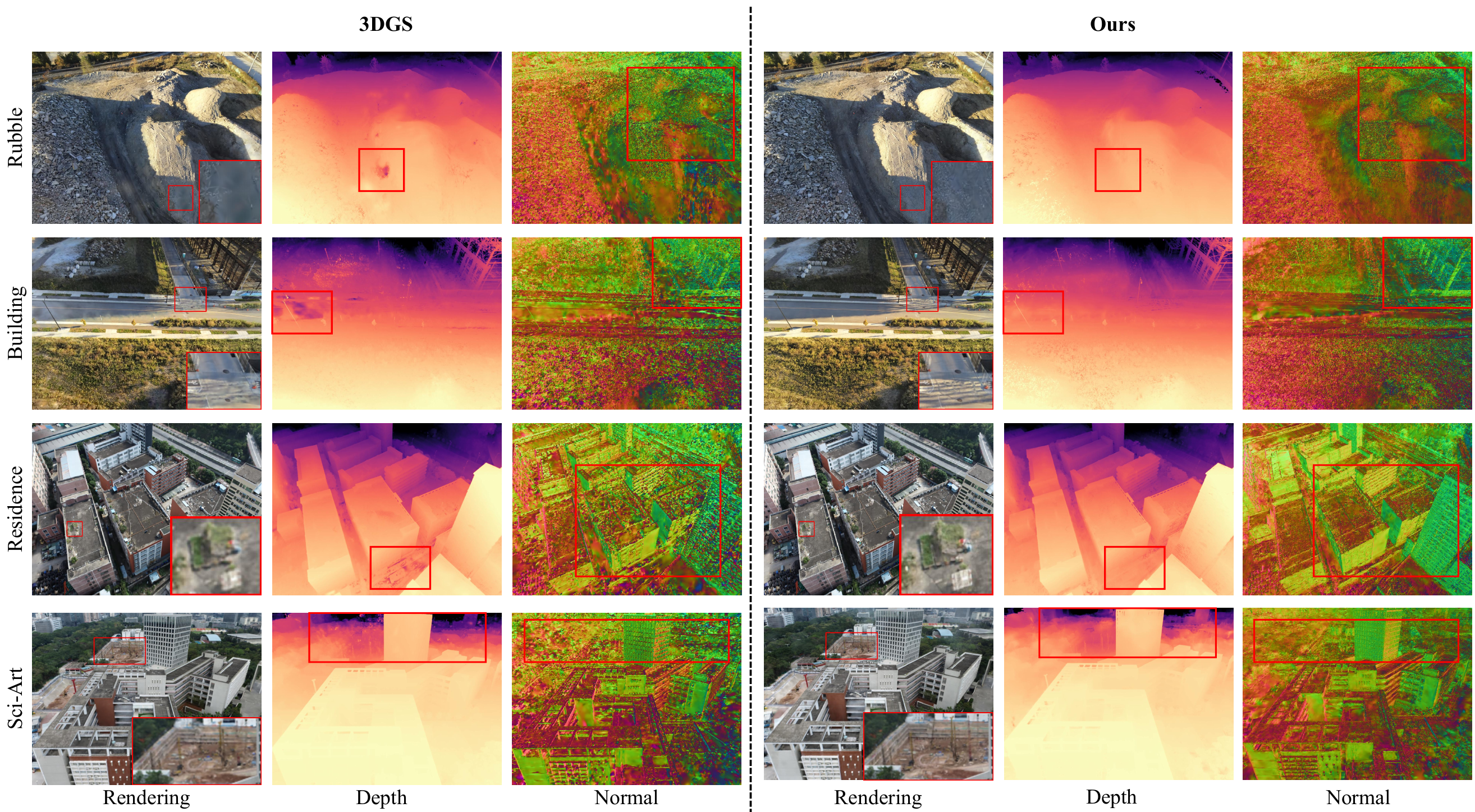}
\caption{Comparison of scene rendering and structural information. Our approach offers enhanced scene rendering and more accurate structural information compared to 3DGS. The regularization constraint in our method not only improves the rendering of fine texture details but also results in more complete depth maps and more accurate object normals.}
\label{depth_normal}
\end{figure*}

\begin{figure*}[!t]
\centering
\includegraphics[width=7.1in]{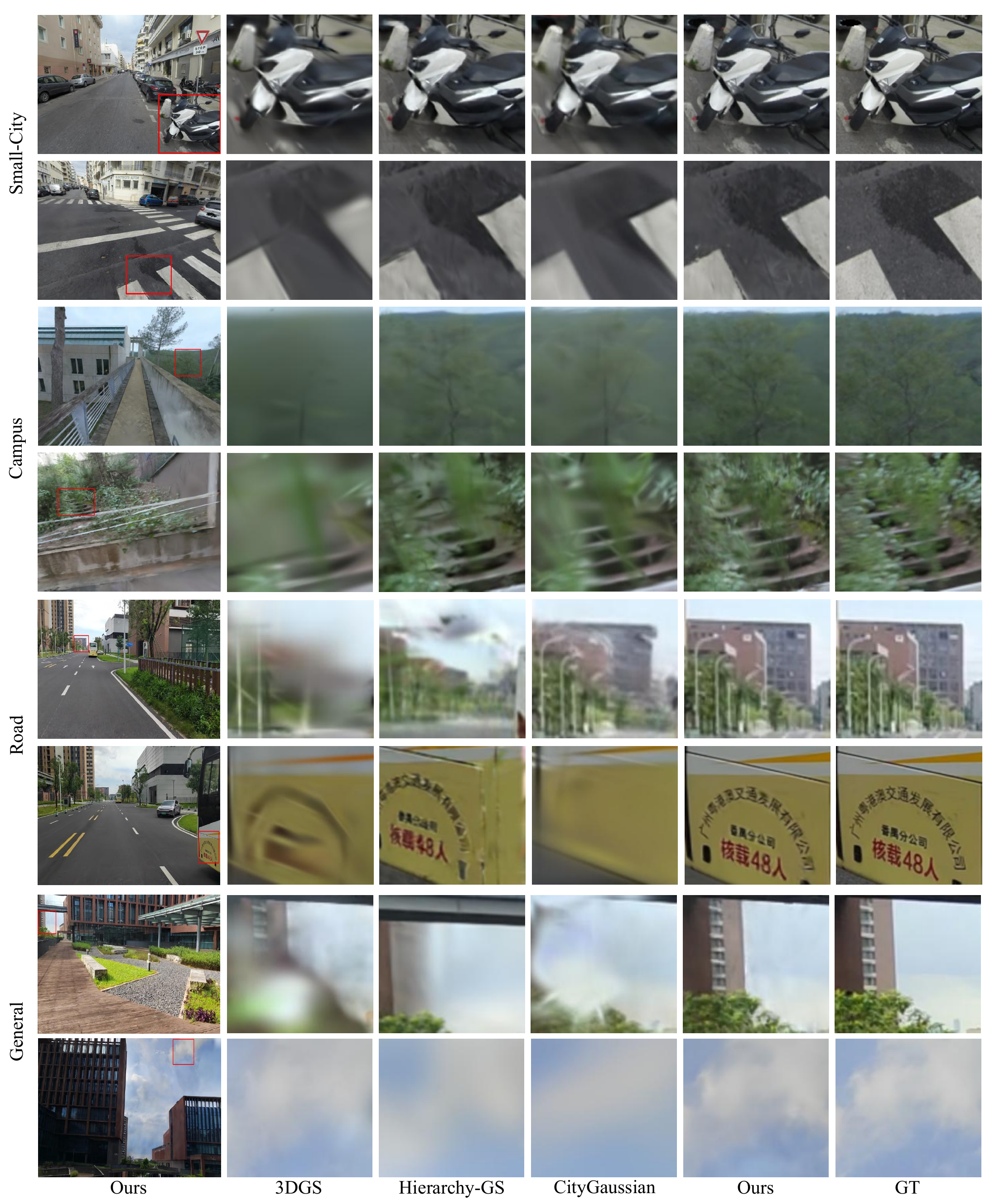}
\caption{Visualization results of four ground datasets. Unbounded scene rendering from a ground-level perspective is inherently challenging. In contrast to existing methodologies, our approach demonstrates superior accuracy in rendering texture details and achieves remarkable clarity in representing far-field elements.}
\label{ground_fig}
\end{figure*}

\begin{table*}[ht]
    \centering
    \caption{Comparative analysis of Gaussian splatting methods across four benchmark datasets. The table presents quantitative results for PSNR, SSIM, and LPIPS. The best and second best results are highlighted with red shadings in bold and yellow shadings with underline, respectively.}
    \scalebox{1}{
    \begin{tabular}{lcccccccccccc}
        \toprule
        Metrics & \multicolumn{3}{c}{Small-City} & \multicolumn{3}{c}{Campus} & \multicolumn{3}{c}{Road} & \multicolumn{3}{c}{General} \\
        \cmidrule(lr){2-4} \cmidrule(lr){5-7} \cmidrule(lr){8-10} \cmidrule(lr){11-13}
        & PSNR$\uparrow$ & SSIM$\uparrow$ & LPIPS$\downarrow$ & PSNR$\uparrow$ & SSIM$\uparrow$ & LPIPS$\downarrow$ & PSNR$\uparrow$ & SSIM$\uparrow$ & LPIPS$\downarrow$ & PSNR$\uparrow$ & SSIM$\uparrow$ & LPIPS$\downarrow$ \\
        \midrule
        3DGS \cite{kerbl20233d} & 25.34 & 0.776 & 0.337 & 23.87 & 0.785 & 0.378 & 20.18 & 0.679 & 0.397 & 19.53 & 0.641 & 0.411 \\
        
        Hierarchy-GS \cite{kerbl2024hierarchical} & \cellcolor[HTML]{FFFFE0}\underline{26.62} & \cellcolor[HTML]{FFFFE0}\underline{0.820} & \cellcolor[HTML]{FFFFE0}\underline{0.259} & \cellcolor[HTML]{FFCCC9}\textbf{24.61} & \cellcolor[HTML]{FFFFE0}\underline{0.807} & 0.331 & 14.67 & 0.405 & 0.538 & 20.03 & 0.655 & \cellcolor[HTML]{FFFFE0}\underline{0.294} \\
        
        CityGaussian \cite{liu2024citygaussian} & 23.68 & 0.810 & \cellcolor[HTML]{FFFFE0}\underline{0.259} & 22.82 & 0.788 & \cellcolor[HTML]{FFFFE0}\underline{0.322} & \cellcolor[HTML]{FFFFE0}\underline{23.88} & \cellcolor[HTML]{FFFFE0}\underline{0.749} & \cellcolor[HTML]{FFFFE0}\underline{0.288} & \cellcolor[HTML]{FFFFE0}\underline{22.61} & \cellcolor[HTML]{FFFFE0}\underline{0.732} & 0.304 \\
        
        Ours  & \cellcolor[HTML]{FFCCC9}\textbf{27.55} & \cellcolor[HTML]{FFCCC9}\textbf{0.880} & \cellcolor[HTML]{FFCCC9}\textbf{0.170} & \cellcolor[HTML]{FFFFE0}\underline{24.35} & \cellcolor[HTML]{FFCCC9}\textbf{0.816} & \cellcolor[HTML]{FFCCC9}\textbf{0.266} & \cellcolor[HTML]{FFCCC9}\textbf{24.26} & \cellcolor[HTML]{FFCCC9}\textbf{0.769} & \cellcolor[HTML]{FFCCC9}\textbf{0.245} & \cellcolor[HTML]{FFCCC9}\textbf{23.29} & \cellcolor[HTML]{FFCCC9}\textbf{0.771} & \cellcolor[HTML]{FFCCC9}\textbf{0.231} \\
        
        \bottomrule
    \end{tabular}
    }
    \label{ground_tab} 
\end{table*}

\begin{figure*}[!t]
\centering
\includegraphics[width=7in]{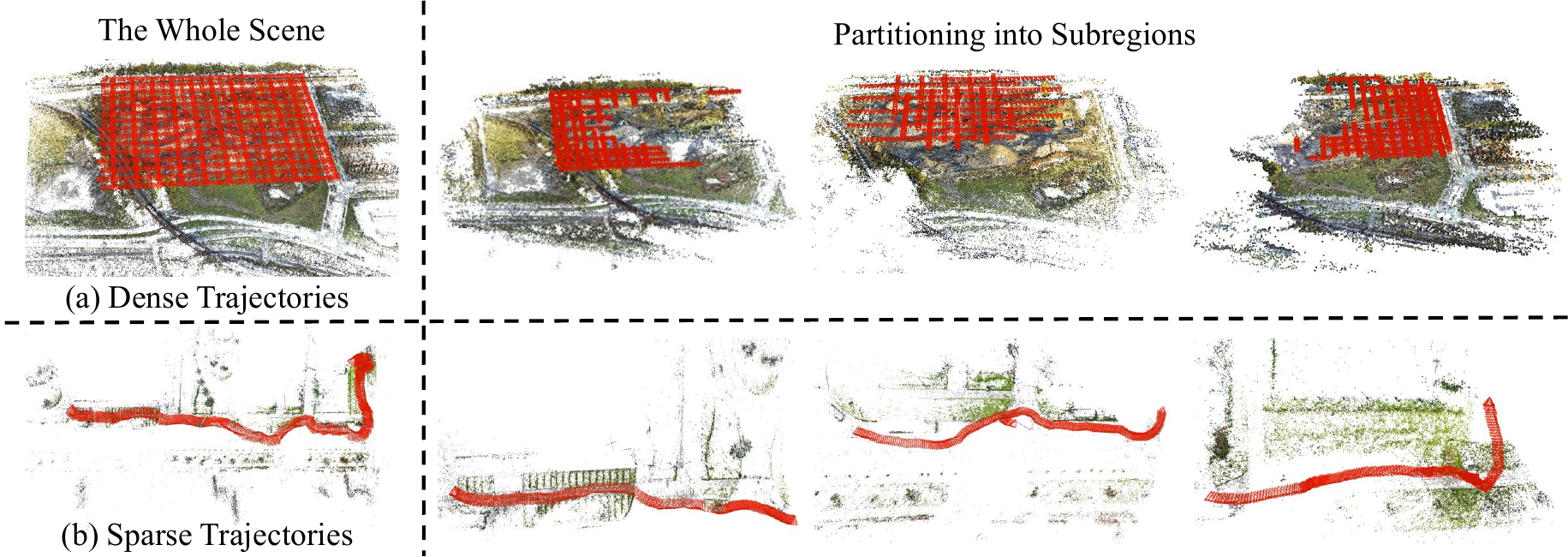}
\caption{Visualization of spatial partitioning based on our graph segmentation. (a) Partitioning cameras and point clouds of large-scale aerial scenes captured by dense camera trajectories into three spatial regions. (b) Partitioning cameras and point clouds of large-scale ground scenes captured by sparse camera trajectories into multiple spatial regions. Our adaptive spatial partitioning method is applicable to different scenes and various camera trajectories.}
\label{region_div}
\end{figure*}

\textbf{Aerial Datasets}. We evaluate our method on four large-scale publicly available aerial imagery datasets: $Rubble$ and $Building$ from the Mill-19 dataset
\cite{turki2022mega}, and $Residence$ and $Sci$-$Art$ from the UrbanScene3D dataset \cite{lin2022capturing}. These four datasets comprise 1,678, 1,941, 2,582, and 3,019 images,  respectively.

\textbf{Ground Datasets}. We evaluate our method on four ground-level datasets: $SmallCity$, $Campus$, $Road$, and $General$. $SmallCity$ and $Campus$ were introduced by \cite{kerbl2024hierarchical} and contain 1,500 and 1,190 images, respectively. $Road$ and $General$ are self-built datasets. $Road$ consists of numerous distant objects and complex street scenes captured with a smartphone along an extended and irregular trajectory. This dataset is designed to validate the robustness of algorithms for distant view reconstruction and textural detail preservation. $General$ comprises smartphone-captured images of double-layer buildings, which aims at validating the algorithms' robustness for spatial partitioning. $Road$ and $General$ contain 647 and 619 images, respectively, with a resolution of 1920×1080. The trajectories for these two datasets are illustrated in Fig. \ref{track}.

\subsection{Implementation Details}

For the four public aerial imagery datasets, we partition each scene into 8 regions and select every 8th image for the test set. The datasets $Rubble$, $Building$, $Residence$, and $Sci$-$Art$ contain 210, 242, 323, and 378 test images, respectively. For the ground-level datasets $SmallCity$ and $Campus$, we employ a single chunk for both training and test processes, which are consistent with the Hierarchy-GS \cite{kerbl2024hierarchical}. For our self-captured datasets, we divide the whole scene into 3 regions and similarly select every 8th image for the test set. The Road and General test sets comprise 81 and 78 test images, respectively. In the multi-view constraint implementation, for each image, we select 4 neighboring cameras with the highest edge weights from the camera connectivity graph. For the multi-scale constraint, we set the scale threshold parameters $g$ and $b$ to 0.01 and 0.1, respectively. Our training consists of 40,000 iterations, with Gaussian pruning applied between iterations 500 and 15,000. The regularization parameter $\lambda$ in the training loss was set to 0.2. In the progressive rendering process, we use a value of 0.5 for parameter $\beta$. All training and test experiments are conducted on a single NVIDIA A6000 GPU with 48GB memory.

\subsection{Results on Aerial Scenes}


A comparison of different methods is shown in Table \ref{aerial_tab}. Due to the spatial structure prior of sparse point clouds, 3DGS-based methods \cite{kerbl20233d} achieve superior rendering performance compared to NeRF-based methods such as Mega-NeRF \cite{turki2022mega} and Switch-NeRF \cite{zhenxing2022switch}. Hierarchy-GS \cite{kerbl2024hierarchical} represents the SOTA approach for ground-level scenes, utilizing maximum densification instead of average densification to enhance fine-grained texture details in close-up views. However, using a simple sky box to model distant views can compromise the reconstruction accuracy of objects far from the principal camera on the ground. Our method employs the graph-structured adaptive spatial partitioning to group cameras with strong co-visibility into the same region. This graph structure further facilitates the identification of neighboring cameras for each camera, enabling the use of multi-view and distant view constraints to improve both texture details and the rendering quality of distant objects. Finally, we replace the combination of VastGaussian \cite{lin2024vastgaussian} and CityGaussian \cite{liu2024citygaussian} with progressive rendering to eliminate floating artifacts caused by Gaussian stacking, thereby enhancing the overall rendering effects. The $Building$ and $Rubble$ datasets exhibit minimal variations in lighting conditions. Under these ideal circumstances, our method achieves superior performance across all metrics. Specifically, it outperforms the SOTA CityGaussian \cite{liu2024citygaussian} by 3.42 dB and 2.59 dB in PSNR on these two datasets, respectively. In contrast, the $Residence$ and $Sci$-$Art$ datasets present significant challenges due to substantial variations in lighting. Despite these difficulties, our method secures the top ranking in PSNR compared to other 3DGS-based techniques, as well as competitive results in both SSIM and LPIPS. Fig. \ref{aerial_fig} presents a visual comparison of the results, demonstrating that our method can vividly render both fine texture details in foliage and windows, as well as distant buildings. Furthermore, both Hierarchy-GS \cite{kerbl2024hierarchical} and CityGaussian \cite{liu2024citygaussian} optimize Gaussian ellipsoids using pixel-wise error based on 3DGS \cite{kerbl20233d}. This lack of structural constraints for the scene leads to errors in depth and normal vector estimation. Fig. \ref{depth_normal} presents a comparison of the rendering results and scene geometry. The images rendered by 3DGS \cite{kerbl20233d} exhibit several blurry regions, and the depth maps contain holes, indicating errors in the object normals. In contrast, our method optimizes 3DGS by incorporating multi-view constraints and multi-scale Gaussian regularization, which leads to clearer rendering of texture details, more accurate depth estimation, and a more distinct representation of surface normals.

\begin{table}[t]
\center
\caption{Ablation studies of different components on the Road dataset, which was partitioned into 3 regions. PR is the Progressive Rendering method. MV is the Multi-View constraint. MS is the Multi-Scale Gaussians.}
\begin{tabular}{|c| c c c | c c c|}
\hline & PR & MV & MS & PSNR $\uparrow$ & SSIM $\uparrow$ & LPIPS  $\downarrow$ \\
\hline (a) & & & & 19.90 & 0.672 & 0.346\\
(b) & $\checkmark$ & & & 23.78 & 0.753 & 0.273\\
(c) & $\checkmark$ & $\checkmark$ & & 23.87 & 0.761 & 0.258\\
(d) & $\checkmark$ & $\checkmark$ & $\checkmark$ & \textbf{24.72} & \textbf{0.775} & \textbf{0.195}\\
\hline
\end{tabular}
\label{table_abla1}
\end{table}

\begin{figure}[!t]
\centering
\includegraphics[width=3.45in]{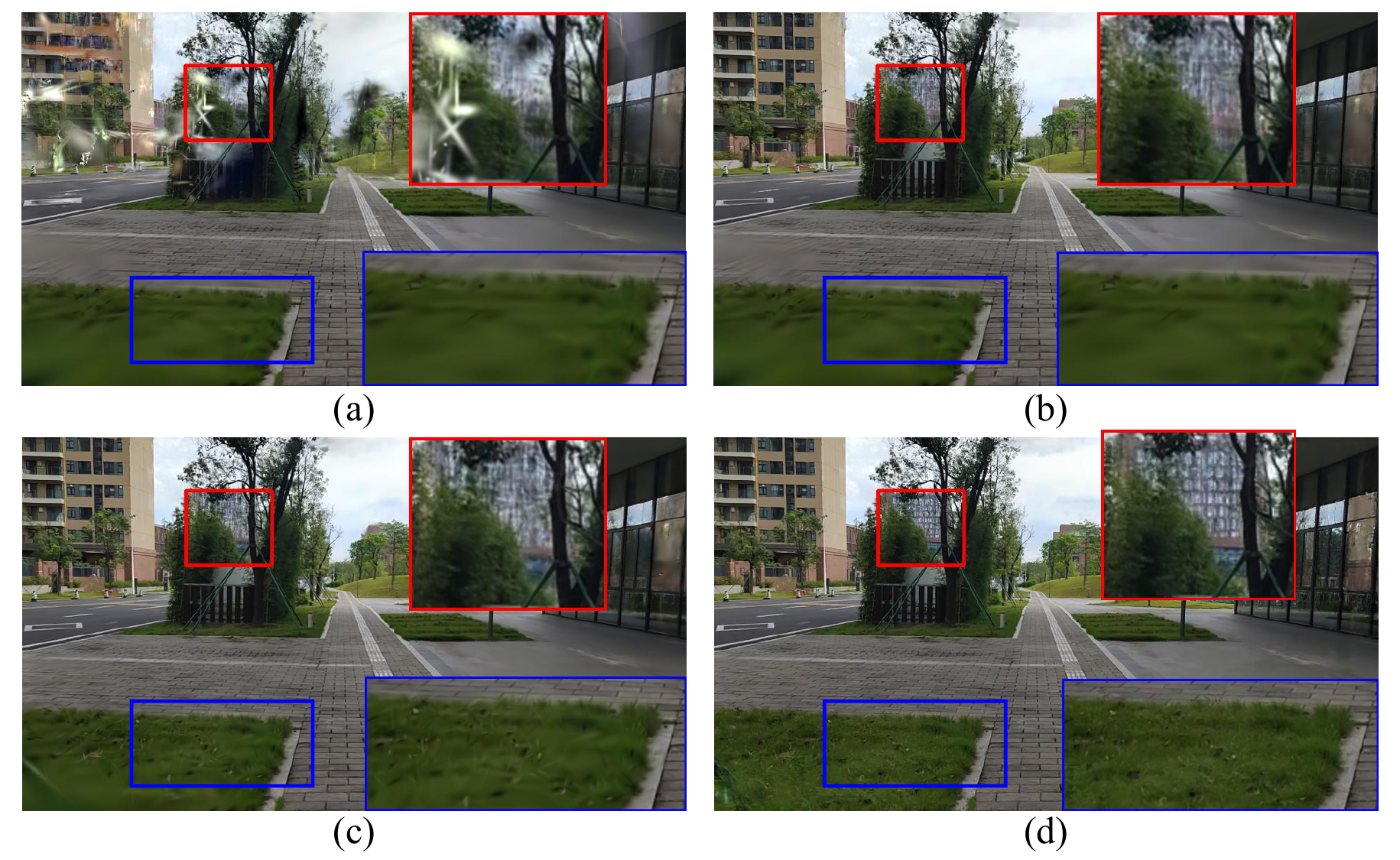}
\caption{Visualization results of our ablation experiments. Comparing (a) with (b) demonstrates that the PR module effectively reduces floating points. Further analysis between (b) and (c) reveals that the MV module significantly enhances the rendering of fine texture details. Finally, the comparison between (c) and (d) illustrates the superior performance of the MS approach in accurately rendering distant scenes.}
\label{abla1}
\end{figure}

\subsection{Results on Ground Scenes}

Comparison experiments in ground-level datasets are presented in Table \ref{ground_tab}. $SmallCity$ and $Campus$ are ground scenes captured with multiple camera trajectories, on which Hierarchy-GS \cite{kerbl2024hierarchical} demonstrates favorable rendering results. In contrast, our own datasets, $Road$ and $General$, consist of ground scenes captured with a single camera trajectory. The point cloud-based simple spatial partitioning of Hierarchy-GS \cite{kerbl2024hierarchical} proves inadequate for these datasets, significantly impacting its rendering performance. Furthermore, CityGaussian \cite{liu2024citygaussian} exhibits limitations in representing distant views, leading to a reduced rendering quality on ground datasets. Our proposed graph-based adaptive spatial partitioning strategy applies to arbitrarily large-scale scenes. The regularization constraints and progressive rendering employed by our method ensure high-quality rendering across all scenes and camera trajectories. The visual comparison results are shown in Fig. \ref{ground_fig}, demonstrating our method's capability to render both texture details and distant views with high fidelity, even vividly capturing the clouds in the sky.

\subsection{Ablation Studies}

\begin{table}[t]
\center
\caption{The relationship between the number of regions, rendering quality, and training time.}
\begin{tabular}{|c| c | c c c c|}
\hline & Regions & PSNR $\uparrow$ & SSIM $\uparrow$ & LPIPS  $\downarrow$ & Time $\downarrow$ \\
\hline (a) & 2 & 25.81 & 0.851 & 0.205 & 1h2m\\
(b) & 4 & 26.95 & 0.865 & 0.184 & 56min\\
(c) & 8 & 28.36 & 0.881 & 0.159 & 52min\\
(d) & 16 & \textbf{28.96} & \textbf{0.882} & \textbf{0.154} & \textbf{47min}\\
\hline
\end{tabular}
\label{table_abla2}
\end{table}

\begin{figure}[!t]
\centering
\includegraphics[width=3.4in]{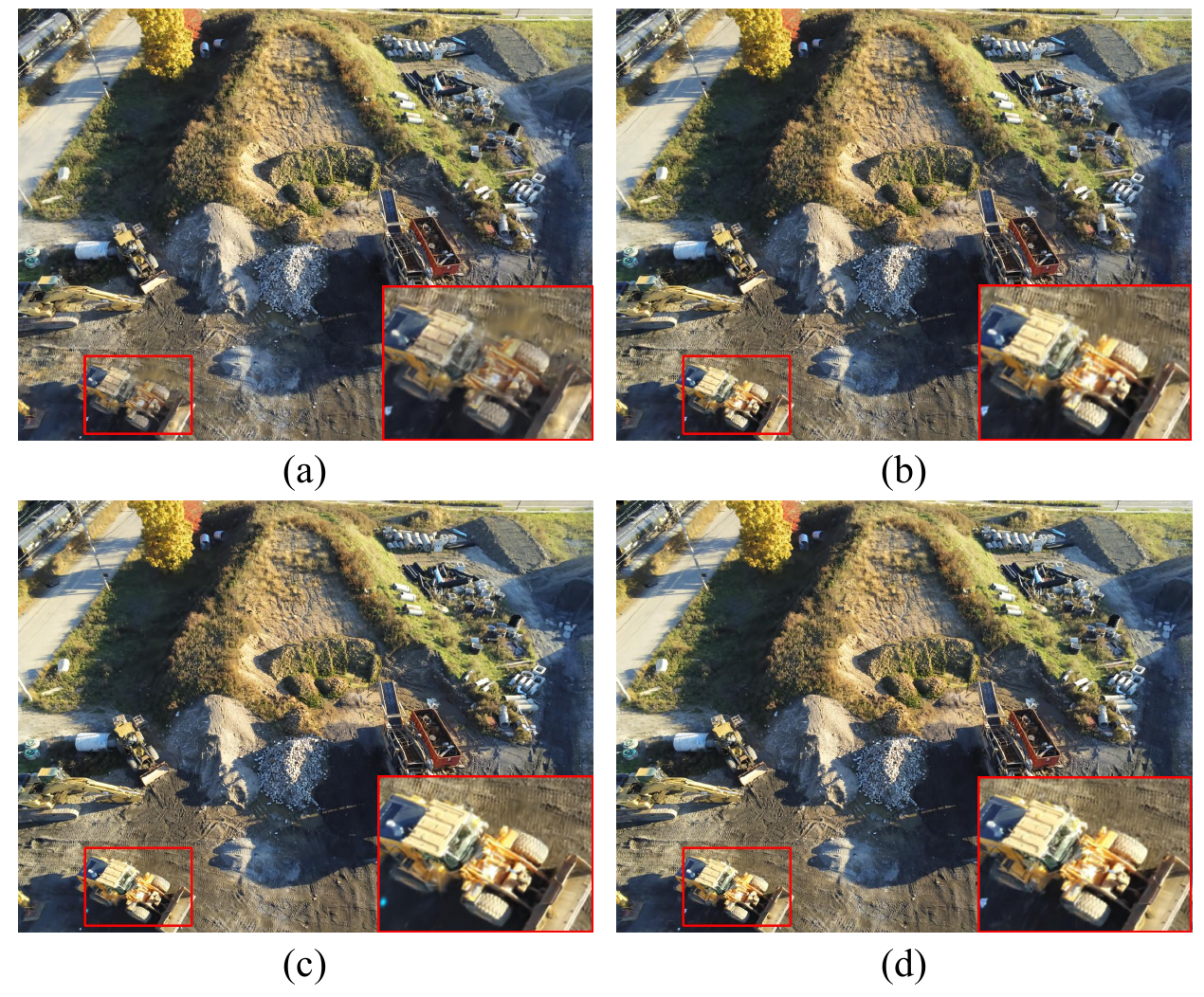}
\caption{Visualization results of rendering with different numbers of regions. As the number of regions increases, the scope of each region decreases, enabling our method to capture more texture details and enhance the rendering quality.}
\label{abla2}
\end{figure}

In this section, a series of ablation experiments are conducted to validate the effectiveness of each module in our proposed method. Our method demonstrates excellent performance across multiple datasets, including both aerial and ground-based scenes, indicating that our graph-based spatial partitioning method can adapt to arbitrarily large-scale scenes and arbitrary camera trajectories, as shown in Fig. \ref{region_div}. Table \ref{table_abla1} quantitatively illustrates the contribution of each component toward enhancing rendering quality, while Fig. \ref{abla1} presents the corresponding visualization results. All experiments were conducted with the scene divided into three regions. Configuration (a) represents the baseline approach, where we train 3DGS \cite{kerbl20233d} independently for each region and then merge local Gaussians into global Gaussians to synthesize novel views, resulting in visible floating artifacts. Configuration (b) builds upon (a) by incorporating the Progressive Rendering (PR) module, which effectively reduces the occurrence of floating artifacts. Configuration (c) extends (b) by adding the Multi-View constraint (MV) module, which significantly improves the rendering quality of textural details, as evidenced in Fig. \ref{abla1}, where the grass rendering shows marked enhancement. Finally, configuration (d) augments (c) with the Multi-Scale Gaussians (MS) module, which substantially improves the reconstruction quality of distant objects. This improvement is clearly demonstrated in Fig. \ref{abla1}, where the building in the distance rendering exhibits significantly enhanced detail. Overall, (d) achieves improvements of 4.82, 0.103, and 0.151 dB in PSNR, SSIM, and LPIPS compared to (a), demonstrating that our method significantly enhances the scene rendering quality.

Furthermore, we verify the relationship between the number of regions and both the rendering quality and the training time per region, as shown in Table \ref{table_abla2}, with the corresponding visualization results presented in Fig. \ref{abla2}. As the number of regions increases, the scope of each region decreases, allowing our method to focus on more texture details, thereby enhancing the rendering quality. Additionally, both the number of images and the point cloud range per region are reduced, resulting in decreased training duration. Compared to configuration (a), configuration (d) demonstrates improvements of 3.15, 0.031, and 0.051 dB in PSNR, SSIM, and LPIPS, respectively, while also reducing the training time by 15 minutes.

\section{Conclusion}

In this paper, we propose TraGraph-GS, a novel view synthesis method based on trajectory graphs that is capable of high-quality rendering for arbitrarily large-scale scenes. Firstly, we design a graph-based spatial partitioning approach that enables our method to be applied to arbitrarily large-scale scenes. Furthermore, a regularization constraint based on the graph structure is developed to enhance texture details and improve the rendering quality of distant objects. We also introduce a progressive rendering strategy to eliminate floating artifacts caused by Gaussian splatting. The proposed TraGraph-GS achieves high-precision novel view synthesis and simultaneously generates high-quality depth and normal maps, which are highly beneficial for scene measurement and mesh reconstruction. As far as limitations, similar to the prior works \cite{liu2024citygaussian} \cite{kerbl2024hierarchical}, our method is influenced by illumination variations. Substantial changes in lighting conditions across the dataset adversely affect the rendering quality. To address these challenging scenarios, future work will focus on developing robust solutions for illumination handling. We plan to investigate appearance encoding techniques to explicitly model illumination effects and decouple lighting information from the input images, thereby reducing inter-image illumination inconsistencies. 



\bibliographystyle{IEEEtran}
\bibliography{IEEEabrv,ref}

\vspace{-10 mm}

\begin{IEEEbiography}
  [{\includegraphics[width=1in,height=1.25in,clip,keepaspectratio]{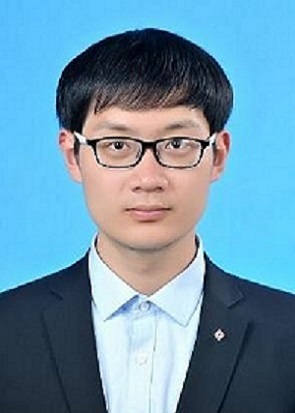}}]{Xiaohan Zhang} received the M.Sc. degree from the School of Mathematics and
  Statistics, Shandong University, Weihai, China, and received the B.S. degree from Xinjiang University, Urumqi, China. He is currently pursuing a Ph.D degree at the School of Future Technology, South China University of Technology. His major is Electronic Information. His research interest is 3D vision, including multi-view stereo, NeRF, 3DGS, and the application of 3D reconstruction technology in large-scale scenes. 
\end{IEEEbiography}  

\vspace{-10 mm}

\begin{IEEEbiography}
  [{\includegraphics[width=1in,height=1.25in,clip,keepaspectratio]{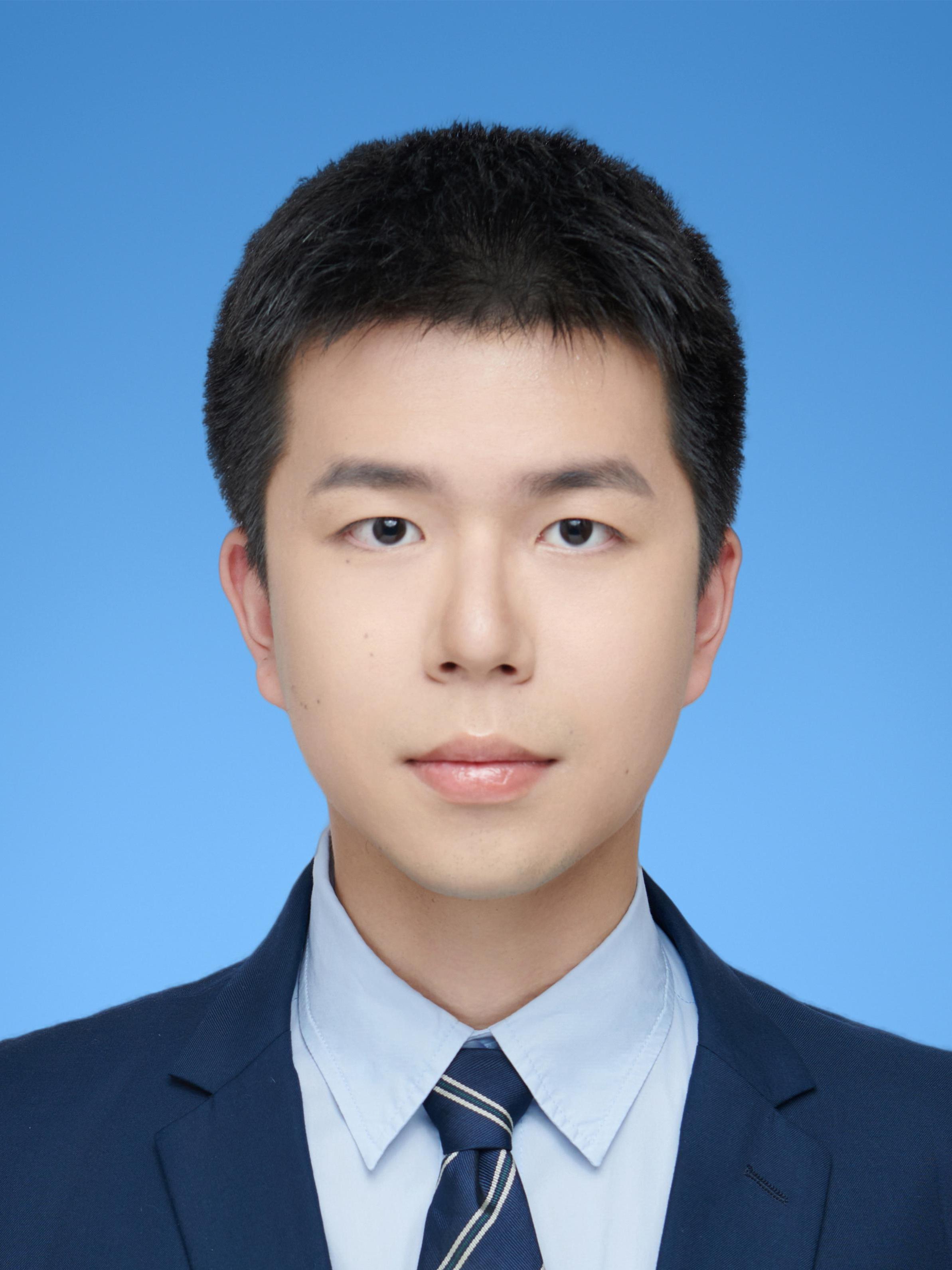}}]{Sitong Wang} is currently pursuing a Bachelor’s degree majoring in Artificial Intelligence at School of Future Technology, South China University of Technology. His research interests include computer vision and computer graphics. 
\end{IEEEbiography}  

\vspace{-10 mm}

\begin{IEEEbiography}
  [{\includegraphics[width=1in,height=1.25in,clip,keepaspectratio]{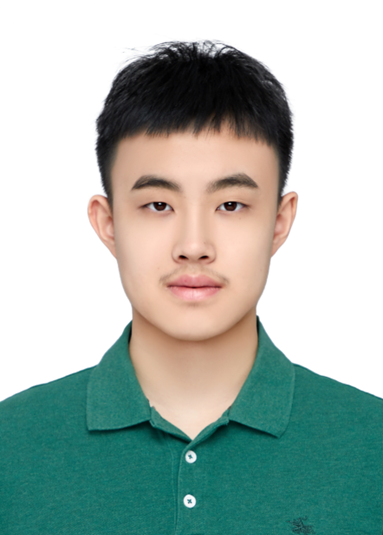}}]{Yushen Yan} is currently pursuing a Bachelor's degree at the South China University of Technology (SCUT). His research focuses on 3D vision, particularly 3D reconstruction technologies.
\end{IEEEbiography}

\vspace{-10 mm}

\begin{IEEEbiography}
  [{\includegraphics[width=1in,height=1.25in,clip,keepaspectratio]{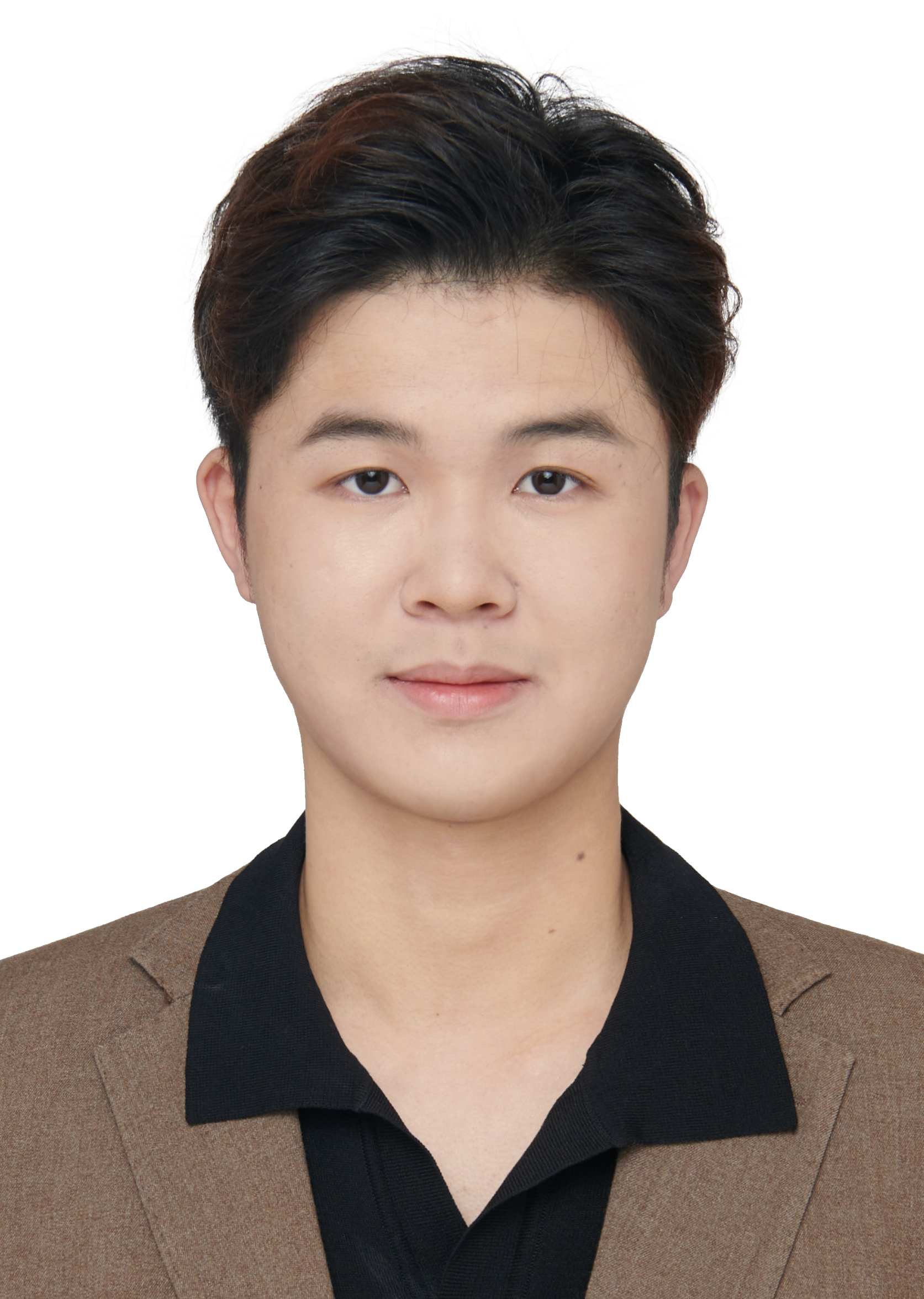}}]{Yi Yang} is currently pursuing a Master's degree at the School of Future Technology, South China University of Technology (SCUT), China. His main research interests include 3D vision, text-to-music generation and text-guided music editing.
\end{IEEEbiography}

\vspace{-10 mm}

\begin{IEEEbiography}
  [{\includegraphics[width=1in,height=1.25in,clip,keepaspectratio]{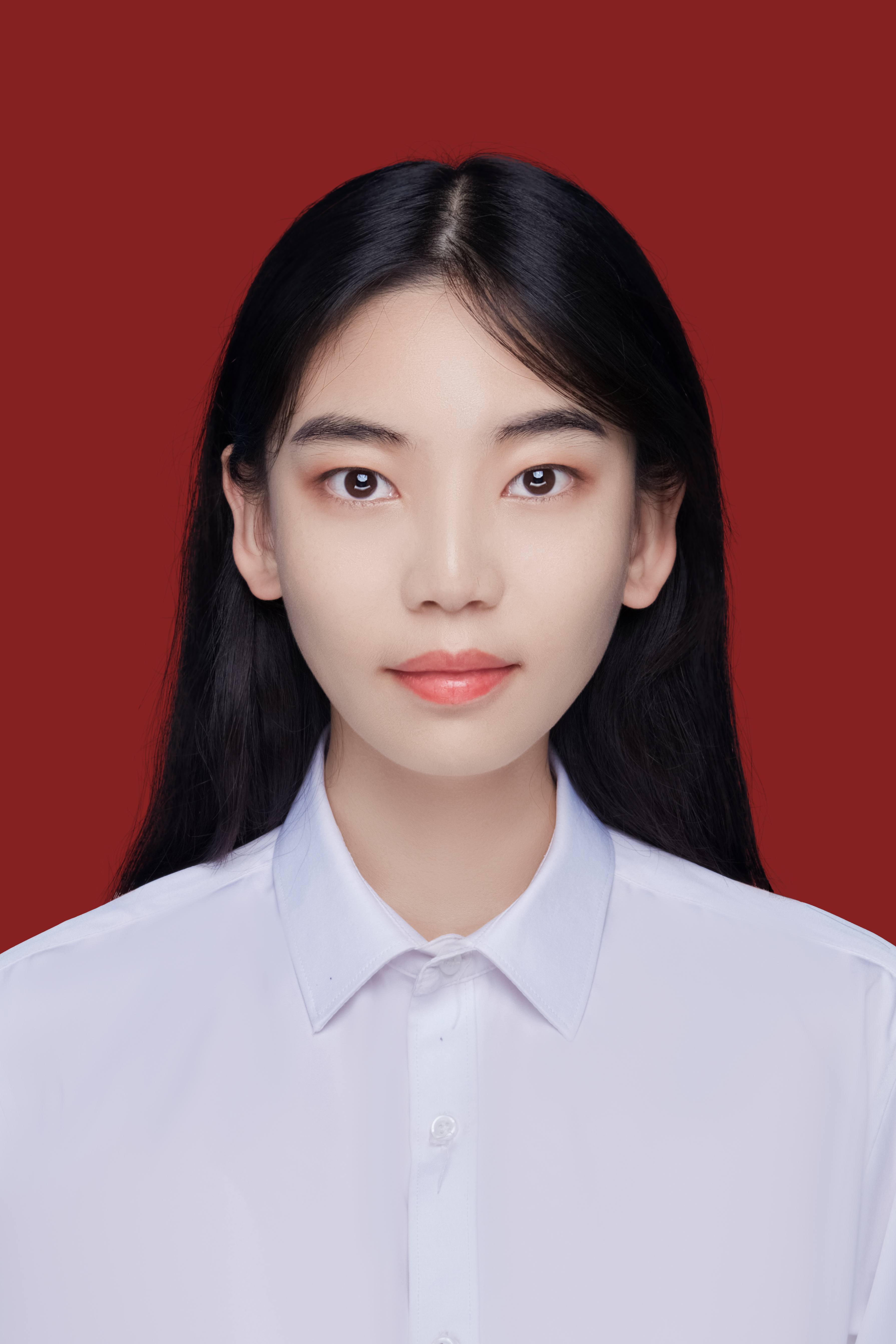}}]{Mingda Xu} received the M.Sc. degree from the Zhongtai Securities Institute for Financial Studies, Shandong University, China, and received the B.S. degree from Xinjiang University, Urumqi, China. She is currently pursuing a Ph.D degree at the School of Future Technology, South China University of Technology. Her major is Electronic Information. Her research interest is 3D vision, including NeRF, 3DGS, and the application of 3D reconstruction technology in different scenes.
\end{IEEEbiography}

\vspace{-10 mm}


\begin{IEEEbiography}
  [{\includegraphics[width=1in,height=1.25in,clip,keepaspectratio]{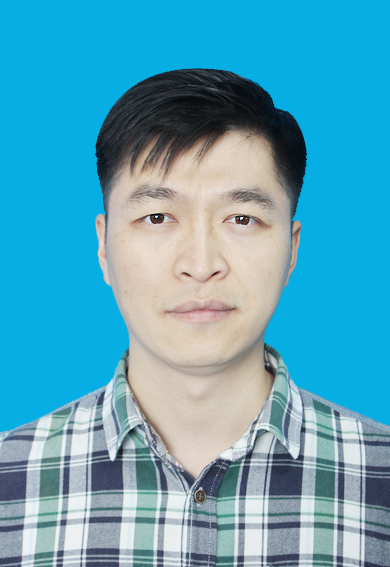}}]{Qi Liu} is currently a Professor with the School of Future Technology at South China University of Technology. Dr. Liu received a Ph.D degree in Electrical Engineering from City University of Hong Kong, Hong Kong, China, in 2019. During 2018 - 2019, he was a Visiting Scholar at the University of California Davis, CA, USA. From 2019 to 2022, he worked as a Research Fellow in the Department of Electrical and Computer Engineering, National University of Singapore, Singapore. His research interests include human-object interaction, AIGC, 3D scene reconstruction, and affective computing, etc. Dr. Liu has published over 50 papers in peer-reviewed journals and conferences, including IEEE TFS, TCYB, TCSVT, IoT-J, ACL, CVPR, AAAI, etc., and has been an Associate Editor of the IEEE Systems Journal (2022-), and Digital Signal Processing (2022-). He was also Guest Editor for the IEEE Transactions on Consumer Electronics, IEEE Internet of Things Journal, IET Signal Processing, etc. He received the Best Paper Award from IEEE ICSIDP in 2019 and ICCBD+AI in 2024.
\end{IEEEbiography}  

\vfill

\end{document}